\documentclass[journal]{IEEEtran}
\usepackage{framed,multirow}
\usepackage{amssymb}
\usepackage{latexsym}
\usepackage{amsmath,graphicx}
\usepackage{xcolor}

\ifCLASSINFOpdf

\else

\fi

\begin{document}
%
\title{Weakly-supervised Learning For Catheter Segmentation in 3D Frustum Ultrasound}
%
%
%

\author{Hongxu Yang, Caifeng Shan, Alexander F. Kolen, Peter H. N. de With
\thanks{Hongxu Yang (h.yang@tue.nl) and Peter H. N. de With are with the Department
of Electrical Engineering, Eindhoven University of Technology, Eindhoven, The Netherlands.}
\thanks{Caifeng Shan (caifeng.shan@gmail.com) is with Shandong University of Science and Technology, Qingdao, China.}
\thanks{Alexander F. Kolen is with Philips Research, Eindhoven, The Netherlands.}}
\maketitle

\maketitle


\begin{abstract}
Accurate and efficient catheter segmentation in 3D ultrasound (US) is essential for cardiac intervention. Currently, the state-of-the-art segmentation algorithms are based on convolutional neural networks (CNNs), which achieved remarkable performances in a standard Cartesian volumetric data. Nevertheless, these approaches suffer the challenges of low efficiency and GPU unfriendly image size. Therefore, such difficulties and expensive hardware requirements become a bottleneck to build accurate and efficient segmentation models for real clinical application. In this paper, we propose a novel Frustum ultrasound based catheter segmentation method. Specifically, Frustum ultrasound is a polar coordinate based image, which includes same information of standard Cartesian image but has much smaller size, which overcomes the bottleneck of efficiency than conventional Cartesian images. Nevertheless, the irregular and deformed Frustum images lead to more efforts for accurate voxel-level annotation. To address this limitation, a weakly supervised learning framework is proposed, which only needs 3D bounding box annotations overlaying the region-of-interest to training the CNNs.  Although the bounding box annotation includes noise and inaccurate annotation to mislead to model, it is addressed by the proposed pseudo label generated scheme. The labels of training voxels are generated by incorporating class activation maps with line filtering, which is iteratively updated during the training. Our experimental results show the proposed method achieved the state-of-the-art performance with an efficiency of 0.25 second per volume. More crucially, the Frustum image segmentation provides a much faster and cheaper solution for segmentation in 3D US image, which meet the demands of clinical applications
\end{abstract}

\IEEEpeerreviewmaketitle

\section{Introduction}
\IEEEPARstart{A}{dvanced} imaging systems, such as fluoroscopy and ultrasound imaging, are widely utilized in the minimal invasive surgery. Typical operations include biopsies and cardiac intervention, which require careful manipulation and coordinate alignment of the instrument in the patient body. The 3D ultrasound (US) is an attractive modality to guide the instrument during the operation, since its real-time radiation-free imaging capacity with rich spatial information for both instrument and tissue anatomy. Nevertheless, 3D US imaging faces challenges of low image resolution, low image contrast, which requires experienced surgeons to interpret the US volumetric data. Therefore, automatic instrument segmentation in 3D US is desired to facilitate the operation.

Instrument segmentation in 3D US is typically performed as a voxel-wise classification task, which assigns a semantic category to each 3D voxel or region. Deep learning methods \cite{litjens2017survey} has been proposed recently to segment the instrument in 3D US data by voxel-level classification \cite{pourtaherian2018robust,yang2019voi}. However, these methods have limitations like compromised semantic information usage by decomposing the 3D information. To address this issue, patch-based strategies were introduced to improve the 3D information usage \cite{Towards,yang2019transferring}. However, patch-based strategy still limits the image-level semantic information processing, which therefore degrades the information usage than a standard but expensive convolutional semantic segmentation in the whole image \cite{long2015fully}. These voxel-based and patch-based approach also hampers the segmentation efficiency for real-time performance, although a faster segmentation was obtained in recent literature by the coarse-to-fine strategy \cite{zhu20183d,yang2020deep}. Beside the above limitations, a precise voxel-level annotation is necessary for Convolutional Neural Network (CNN) training is expensive and laborious to obtain. 

\begin{figure*}[htbp]
\centering
\includegraphics[width=16cm]{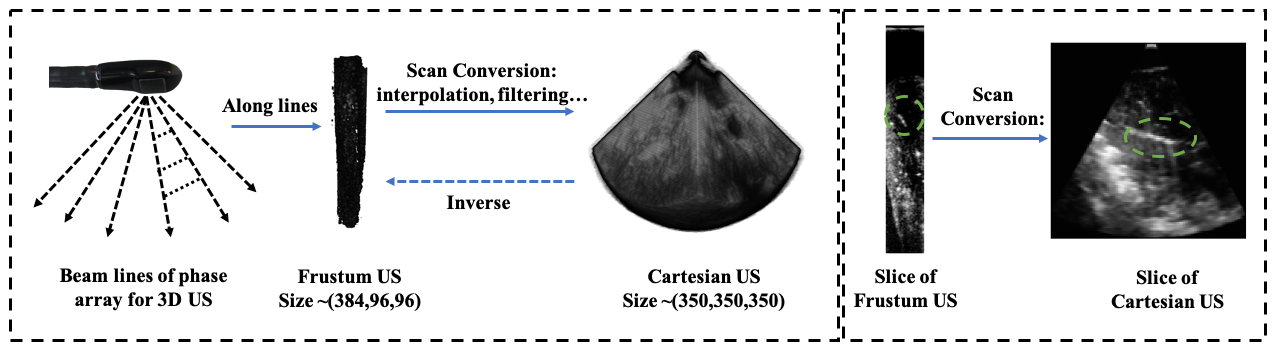}
\caption{Examples of Frustum US and Cartesian US volumes. Because of the beam lines from phase array, the spatial resolution is getting lower as the depth increasing, which is denoted as dot lines between the beam lines. The Frustum image is constructed based on the beam lines, which is then converted into Cartesian image with known angle and spatial information. The standard volume sizes are shown for Frustum and Cartesian US images. Green circles denote the catheter.}
\label{example}
\end{figure*}

The existing methods, including handcrafted feature designing based method and deep learning methods, are all performing segmentations in a Cartesian US images, which was produced by scan conversion and interpolation. The converted Cartesian US images typically have much larger volume size than the one before the scan conversion, so-called the Frustum US, which is the main bottleneck for segmentation efficiency improvement. The relationship between Frustum US and Cartesian US is shown in Figure.~\ref{example}. Comparing to Cartesian US, Frustum US provides advantages of smaller volume size (around 1/7 of the Cartesian image), same spatial information and without extra filtering or smoothing during the scan conversion. More crucially, Frustum format is a standard data type for US data storage and communication between different devices, due to the limited network bandwidth. In this paper, we propose to segment the instrument in Frustum US, which improves the computation efficiency and hardware requirement. To the best of our knowledge, this article is the first one to exploit Frustum US for instrument detection. 

Because of the nature of imaging reconstruction and sound propagation, the obtained Frustum images have irregular information for both instrument and tissue, which leads challenges for voxel-level annotation. Although the accurate annotation can be obtained by inverting of the scan conversion, it is still expensive to obtain the voxel-level annotation.  In this work, we address the above challenges by proposing a novel CNN-based weakly supervised learning method, which learns to segment the catheter in Frustum domain using training data with bounding box (Bbox) annotation. Specifically, by exploiting the information from Bbox annotation, pseudo labels for training images are generated under the guidance of the Class Attention Maps (CAMs) \cite{zhou2016learning} and line filtering \cite{frangi1998multiscale}. With CAMs learned by CNN, the catheter location estimation is obtained with probability-like response, which indicates the most possible area containing the catheter. However, CAMs cannot be directly utilized as supervision signal for segmentation due to its limited resolution and inaccurate boundary estimation. Although CAMs are commonly adopted with Conditional Random Field (CRF) to generate pseudo labels for RGB images \cite{krahenbuhl2011efficient, huang2018weakly}, it is difficult to be applied in gray US images, as the object's edge has lower contrast than the nature images. In contrast, line filtering method, so-called Frangi vesselness filter, has been proven to be a useful spatial descriptor to capture the tubular like structure in 3D space \cite{uhervcik2013line,pourtaherian2017medical}. Nevertheless, without proper location guidance, line filtering can lead to too many false positive voxels and failed to extract the accurate pseudo labels. Our motivation is that, by exploiting voxel's information in both CAMs (providing catheter's voxels location distribution) and line filtering (providing catheter's shape information distribution), its neighboring voxels' responses and location information are correlated to define the voxel's category by CRF \cite{krahenbuhl2011efficient}. Intuitively, the voxels around the target regions are always carrying similar discriminating formation, since the their semantic labels of the catheter have spatial continuity. Compared to Cartesian US volume, Frustum domain images require less GPU computation and memory, but a full volume end-to-end segmentation is still computational expensive. To further accelerate the inference efficiency, a region-of-interest (ROI) segmentation decoder is proposed to segment the catheter with selected feature map. As a consequence, the overall efficiency is further improved instead of decoding the feature maps on the whole volume. 

In our experiments, the proposed method was extensively studied on the challenging ex-vivo RF-ablation catheter dataset in Frustum domain. The results demonstrate that our method achieved the state-of-the-art performance. In addition, we provide an analysis of the proposed method by carrying out detailed ablation studies. More crucially, our results indicates the fact that the segmentation in Frustum domain could achieve much better performance in Cartesian images while it provides more than 10 times less computation cost and memory usage, which indicates a promising result for clinical practice. Based on the complexity analysis \cite{molchanov2016pruning} of the proposed model, Frustum US image uses 1.8 GFLOPs (giga floating point operations) while it is 23.9 GFLOPs for a Cartesian image. Moreover, with ROI-based decoder, the computation cost is reduced from 5.2 GFLOPs to 1.8 GFLOPs in Frustum image. 

In summary, the main contributions of this article are summarized in three-folds: (1) To the best of our knowledge, this article is the first one to make use of Frustum US images for segmentation, which demonstrates a promising efficiency potential for clinical practice. (2) We propose a novel weakly supervised learning framework for catheter segmentation by exploiting the bounding box information, which exploits intra-volume discriminating information for pseudo label generation. (3) Experimental validation demonstrates the proposed method achieves around 66\% Dice score. Meanwhile, the results of our method is comparable to the fully supervised approach with high quality voxel-level annotation. 

The rest of the article is organized as follow. Related works are discussed and summarized in Section.~\ref{Works}, which briefly summarizes literatures in medical instrument segmentation in 3D images and weakly supervised learning methods. The proposed method is explicitly described in Section.~\ref{Methods}. Experiment set ups and results discussion are expressed in Section.~\ref{Experiments} and \ref{Results}, respectively. Finally, Section.~\ref{Conclusions} provides the conclusions of the paper.

\section{Related Work}\label{Works}
In this section, related literatures in catheter segmentation and weakly supervised learning methods are discussed. We first discuss the literatures related to semantic segmentations in 3D medical images, and then the related literatures exploiting weakly supervised learning are discussed in the rest parts.
\subsection{Instrument segmentation in 3D US images}
Medical instrument in 3D medical images, especially 3D US, has been studied in recent years. Line filtering based method were extensively studied as the spatial descriptor for medical instrument, like needle or catheter, which has been proven to be a feasible solution to extract the tubular-like structures in 3D images \cite{uhervcik2013line,pourtaherian2017medical,zhao2013automatic}. Nevertheless, this line-filtering solution may be failed when the anatomical structures of tissues are too complex, e.g. leads to too much outliers. In contrast, the learning-based methods have been extensively studied in recent years, as they could better model the spatial and intensity information by the data-driven approach \cite{pourtaherian2018robust,arif2019automatic}. As for the medical instrument or organ segmentation, a typical solution is adopting the CNN with U-Net like architecture \cite{ronneberger2015u}, which employs encoder and mirrored decoder for multi-scale information exploration. Although these approaches have shown great successful results for their tasks, there are still some limitations for their architecture. First, as for deep learning methods, accurate voxel-level annotation is necessary to train the network, which is expensive and time consuming. Moreover, accurate annotation in low contrast US images are difficult and may lead to noise. Second, the most of the U-Net based method may wast the computational resources on non-interested regions in 3D images \cite{huang20203}, especially for instrument segmentation task, where the instrument takes around 1/1500 voxels of the whole volume. Therefore, it is optimal with respect to clinical practice. Third, most of the U-Net like networks are based on a customized structure, which need to be train from scratch and therefore the network cannot be trained properly with limited dataset. Although there are some works to address these challenges by adopting pre-trained network with Mask R-CNN in 2D medical images \cite{almubarak2020two}, it is still challenging to implement this architecture in 3D cases, as it requires higher computational resources and hard to train with limited 3D medical image dataset. Different from the above detection network with complex architecture, we aim to improve the segmentation efficiency and to reduce the computation complexity by employing ROI-based segmentation in end-to-end training, which preserves the semantic information in whole image while improving the segmentation efficiency. 

\subsection{Weakly supervised learning in medical images}
As for weakly supervised learning, several solutions have been exploited in recent literatures, such as using bounding boxes and scribbles as the information supervision. Dong \emph{et al.} \cite{dong2019instance} proposed to use 3D Faster R-CNN with peak response map to segment the multiple instances in the brain volumetric images. Nevertheless, in contrast to catheter segmentation in 3D images, multiple instances provide richer discriminative information compared to a single instrument case, which leads to an easier situations for training. As for the Faster R-CNN based method, it requires a large amount of dataset to training the region proposal network, which is however difficult to achieve in our case with limited training images. Wu \emph{et al.} \cite{wu2019weakly} proposed to use CAM as the location cue for a representation model, which is then ensemble multiple segmentation candidates as the segmentation map. The CAMs are generated by exploiting whether the images containing the lesion or not. Moreover, to simplify the the classification network, their network compresses the 3D volumes into 2D images, which however hampers the information usage. To achieve the final segmentation, a simple voxel affinity is computed based on CAMs, which however cannot properly exploit the information as the CAMs only provide an estimation of the location instead of a clear boundary (especially in US other than MRI images). Image-level class labels have been widely exploited in computer vision area \cite{ahn2018learning}, which considers multiple image classes to generate CAMs for further operations, such as pseudo label generations. Nevertheless, this approach requires more images with different instrument classes, which is difficult to achieve by collecting the complex datasets. With above literatures, there are still some limitations for catheter segmentation in 3D US based weakly supervised learning: (1) 3D US, especially Frustum US, has worse image contrast than commonly studied modalities, such as MRI or CT. (2) Limited datasets to train a sufficient network to localize and segment the target, such as Faster R-CNN. (3) Multi-class images collection is not feasible for US-guided intervention. As a result, these methods are challenged when considering catheter segmentation in Frustum US.

CAMs could exploit the target-related information for location, such as B-line localization in B-mode images \cite{van2019localizing}. As a consequence, as shown in the above section, it is commonly applied to generation a coarse location of the target region, which is then refined by dense CRF or affinity analysis for fine segmentation. Nevertheless, due to low resolution and partially segmented probability-like activation maps, as shown in Figure.~\ref{CRF}, it is not possible to obtain an accurate segmentation in challenging ultrasound images when compared to MRI or common RGB images. Alternatively, CAMs can be used to generate pseudo labels with proper processing. Huang \emph{et al.} \cite{huang2018weakly} proposed to use CAMs to generate target region seeds, which is then refined by deep seeded region growing for pseudo label generation. Similarly, Ahn \emph{et al.} \cite{ahn2019weakly} proposed to use CAMs with affinity network to generate labels for weakly supervised learning. Nevertheless, compared to nature images with multiple classes datasets, catheter in 3D Frustum US cannot generate a clear boundary based on CAMs. To address the limitation of the CAMs-based method, we employs the Frangi line filter as complementary information, which exploits boundary information of the catheter, and therefore improves the discriminating information for pseudo labels.

\section{Our Method}\label{Methods}
\begin{figure*}[htbp]
\centering
\includegraphics[width=16cm]{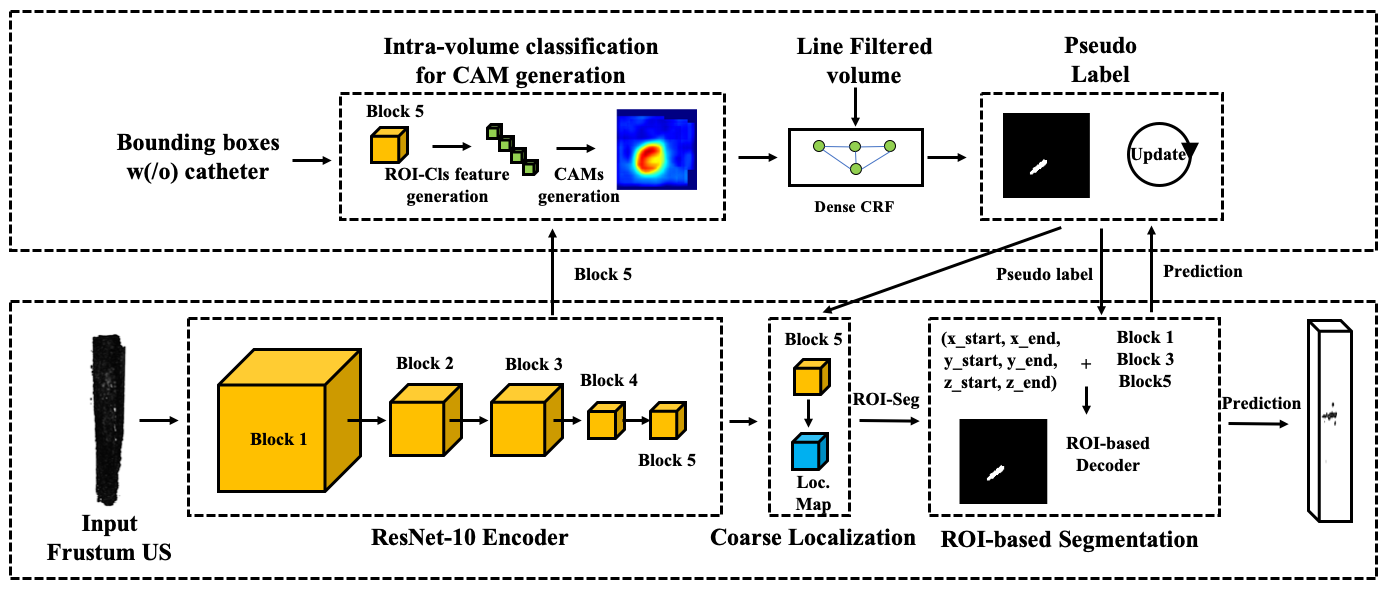}
\caption{A schematic illustration of the proposed weakly supervised learning framework with CAM-guided label generation. The method includes two main branches based on the ResNet encoder: (1) pseudo label generation based on the proposed framework, and (2) catheter segmentation based on coarse localization and regional decoding.}
\label{overivew}
\end{figure*}

The overview of the proposed network is shown in Figure.~\ref{overivew}. Specifically, it includes two modules for different tasks: (1) Intra-volume classification for input Frustum volume, which is to distinguish whether the selected regions including catheter or not. The classification network is used to generated the CAMs, which is compounded with line filtering response to generate pseudo annotation. (2) The catheter localization and segmentation are applied sequentially, which are trained based on the pseudo annotation from the first step. During the training, the classification branch learns the regional discriminative information of the catheter, and generates the CAMs maps for pseudo labels, which compounds the response map from line filtering under the guidance of the dense CRF \cite{krahenbuhl2011efficient}. Then, these pseudo labels are used to train the localization branch to extract the spatial regions containing the catheter, which is processed by the decoder for semantic segmentation. The pseudo labels are generated on-the-fly that the network could gradually interactive within CAMs, pseudo labels and predictions. As a result, the the proposed approach gradually exploit the discriminative information within the provided bounding box. More details are introduced as follow.

\subsection{CAMs-guided pseudo label generation}
The CAMs-guided label generation consists of three different steps: (1) CAMs generation from intra-volume classification network, (2) initial pseudo label generation based on CAMs, line filtering and dense CRF, and (3) iteratively refinement for pseudo label based on history and predictions during the segmentation network training. More details are discussed as follow.

\begin{figure}[htbp]
\centering
\includegraphics[width=8cm]{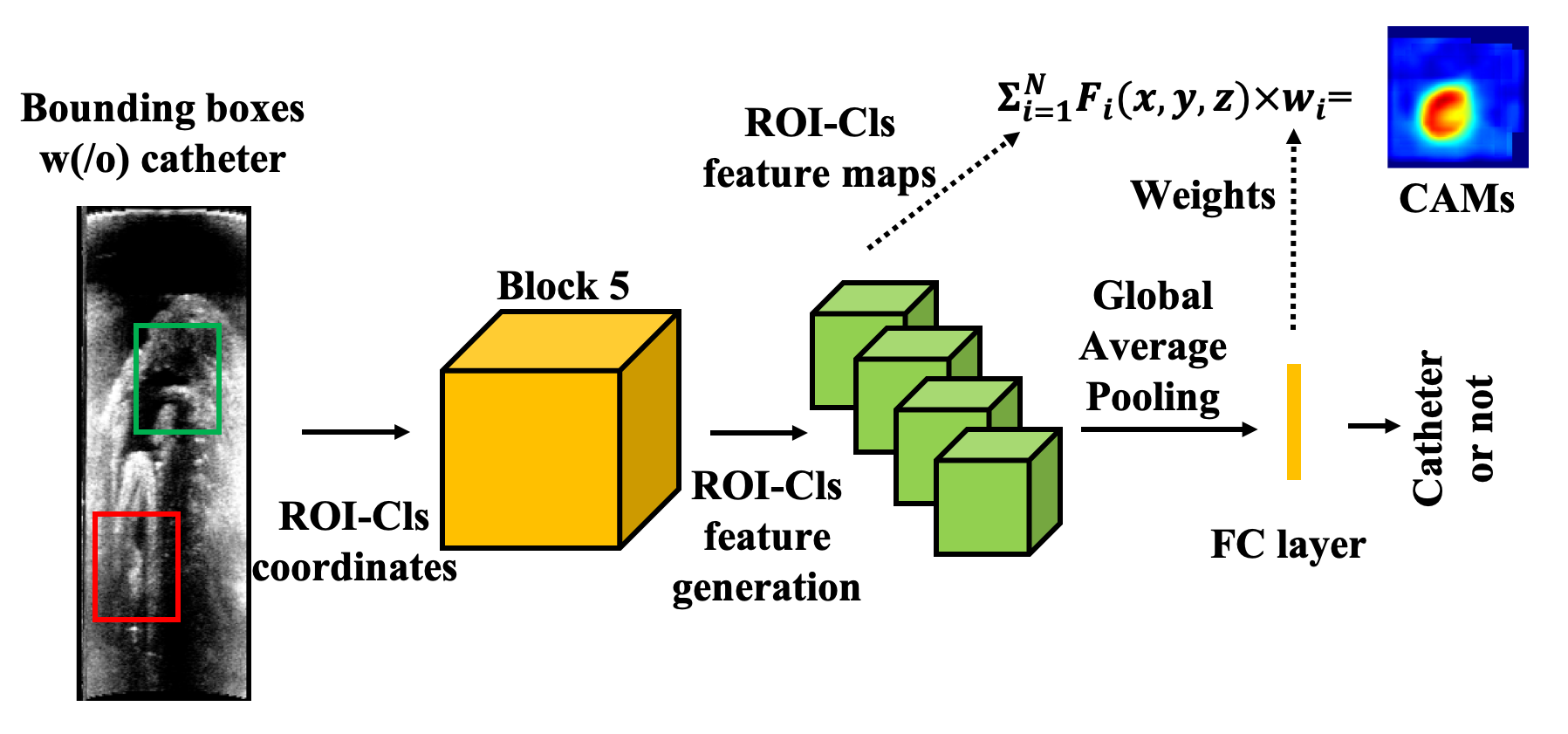}
\caption{The probability map is generated by compounding CAMs, line filtering response and input volume. By exploiting the spatial correlation of voxels by CRF, pseudo label is generated. During the training, Pseudo labels are updated by incorporating with predictions from the model. Green box is catheter, and red box is non-catheter.}
\label{CAMs}
\end{figure}

\noindent {\bf Intra-volume classification for CAMs generation}: We proposed an intra-volume classification framework, which is shown in Figure.~\ref{CAMs}. To generate CAMs, a common practice is to include two different image types, i.e. containing target or not, at image-level. However, it requires extra dataset collection and annotation effort. Moreover, when the catheter occupies extremely small space compared to the background tissue, we failed to achieve a successful classification results in our application. 

In our proposed classification method, the input volume is encoded by the backbone network, i.e. ResNet-10 in this article, because of its compact and satisfied performance architecture \cite{chen2019med3d}. Strides of all the layers are set to be 1 except block 2 to keep the spatial resolution (stride=4 in total). Because of the bounding box annotation, feature maps' regions with and without catheter can be randomly extracted as the input for the classification module. With the extracted feature maps, global average pooling is applied to reduce the spatial dimensions and is feed into a fully connected layer (FC layer with 512-2 nodes). The prediction gives the probability of the input regions are catheter or not.

With a converged classifier, the fully connected layers' neuron weights w.r.t. feature maps can be extracted as the feature map weight for the Class Activation Maps generation \cite{zhou2016learning}. For each feature map $F(x,y,z)$ at location $(x,y,z)$, its CAMs are generated as 
\begin{equation}
\text{CAMs}(x,y,z)=\sum_{i=1}^{N}w_i\times{F_i(x,y,z)},
\end{equation}
where $w_i$ is the weight from FC layer for feature map $F_i(x,y,z)$. Then the obtained CAMs is normalized into range $[0,1]$ as the probability map. Example CAMs is shown in Figure.~\ref{CAMs}. By comparing the CAMs with US image, CAMs only provides a coarse catheter location in the 3D volume, therefore, it is not sufficient for pseudo label generation directly.

\begin{figure}[htbp]
\centering
\includegraphics[width=6cm]{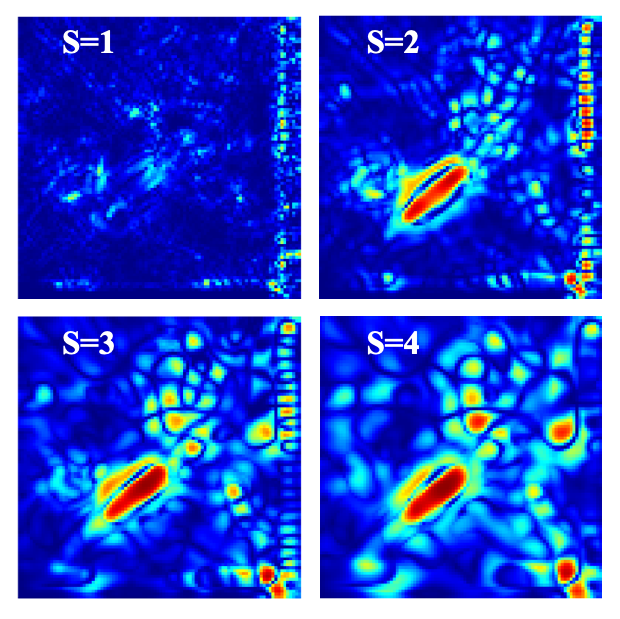}
\caption{Four different kernel value for Gaussian filtering in line filtering method. Difference scales extract different vesselness information.}
\label{frangi}
\end{figure}

\noindent {\bf Initial pseudo label generation}: Compared to CAMs, which provides an estimation of the catheter location in 3D space, line filtering, so-called Frangi vesselness filter, could estimate the catheter shape and boundary with a-priori structure knowledge. As is defined by Frangi \emph{et al.} \cite{frangi1998multiscale}, it is a low-high level analysis from voxel filtering to contextual shape. Example response maps for different kernel values are shown in Fig.~\ref{frangi}. As can be observed, tissue regions are also interpreted as vessleness response as the value increase, but multiscale boundary details of the catheter included. In our case, we experientially select a range of value $s\in\{2,3\}$ to extract more details of the catheter and omit the outliers, while the final response is obtained by maximum operation among scales for each voxel. 

\begin{figure}[htbp]
\centering
\includegraphics[width=8cm]{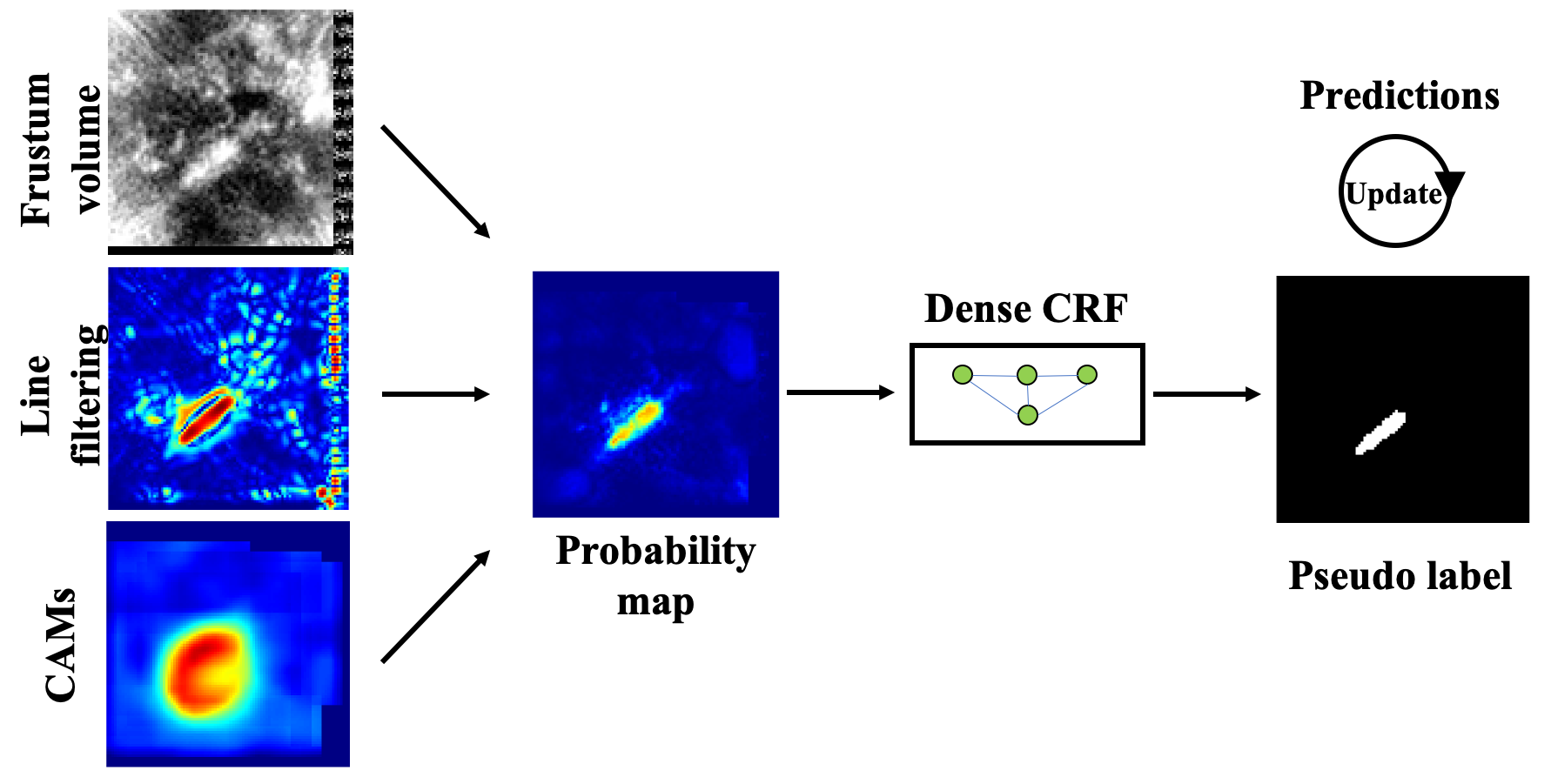}
\caption{The probability map is generated by compounding CAMs, line filtering response and input volume. By exploiting the spatial correlation of voxels by CRF, pseudo label is generated. During the training, Pseudo labels are updated by incorporating with predictions from the model.}
\label{CRF}
\end{figure}

\begin{figure}[htbp]
\centering
\includegraphics[width=8cm]{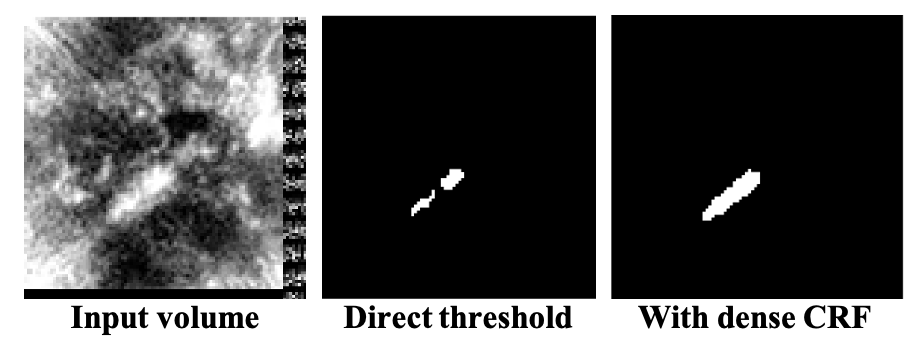}
\caption{Examples of input volume, pseudo mask with direct thresholding and with dense CRF based on $\mathcal{U}$.}
\label{CRFcompare}
\end{figure}

Although these non-catheter regions are included from line filtering, they are compressed by CAMs with multiplication. To preserve more details of the catheter boundary, original volume is also considered to generate the final probability map, which is shown in Figure.~\ref{CRF}. The formula of the final probability map is defined as
 \begin{equation}\label{initial}
\mathcal{U}=\mathcal{V}\times\text{CAMs}\times{I}/255,
\end{equation}
where $I$ is the input volumetric data with intensity range of $[0,255]$. As a result, the final probability-like catheter map in volumetric data is generated. However, as can be observed in Figure.~\ref{CRFcompare}, a straight forward threshold on $\mathcal{U}$ would lead to unsatisfied mask, as it provides inconsistent probability distribution due to the limitations of CAMs. To better exploit the spatial relationship among voxels at the catheter boundary and preserve the catheter's voxels at discontinuity region for a better pseudo label, dense CRF is considered as the post-processing steps. Dense CRF could exploit voxels neighboring points, which potentially sharing a same context. As a consequence, small isolated regions or holes in the generated mask could be further refined as the pseudo label. As shown in the above, the input unary of the dense CRF is obtained by thresholding on $\mathcal{U}$, while its corresponding input image is $I$.

\noindent {\bf Refined pseudo label generation}: Although the initialized pseudo label is obtained based on the above method, it is mostly based on priori knowledge with incorrect CAMs estimation. To better exploit the voxel information from decoder and use the discriminating information, we proposed to iteratively update the pseudo annotation based on segmentation histories. Specifically, for epoch $t$, its corresponding pseudo annotation is denoted as $y_t$, which is defined as 
 \begin{equation}\label{update}
y_t=\text{CRF}(I,\mathcal{U}_{t-1}\times\eta+\hat{y}_t\times\mathcal{U}_{t}\times(1-\eta)),
\end{equation}
where CRF is the dense CRF operation, and $\mathcal{U}_{t}$ is probability map based on Eqn.~\ref{initial}, and $\hat{y}_t$ is the predicted volume at epoch $t$. Parameter $\eta$ is the decay weight to balance the history and current predictions as moving average. Specifically, the initial $\mathcal{U}_0$ is obtained based on Eqn.~(\ref{initial}) without decay weight. By doing so, the pseudo annotation is gradually converging to the predictions from the network while employing the historical information for information stability.
\subsection{Catheter segmentation and training}
Based on the generated pseudo annotation, a coarse localization step is applied on the feature maps from the backbone, which estimate the spatial coordinate of the catheter. With obtained the coordinate, and its corresponding annotation, a finer segment decoder is trained. To better exploit the discriminating information by multi-task learning, a joint training strategy is applied. More details are discussed as follow.

\noindent {\bf ROI-based segmentation}: Based on pseudo masks from CAMs-guided line filtering, a coarse localization is firstly introduced to learn the coarse location of the catheter. Although classification brach could distinguish the categories of the selected ROIs, i.e. region contains catheter or not, this regional method cannot directly applied on the whole volumes efficiently. Alternatively, we construct a simple localization branch to estimate the region of the catheter. Specifically, Block 5 from the encoder is then followed by $1\times1\times1$ kernels to reduce the feature map dimension from 512 to 1 generate a probability map, which indicates the coarse location of the catheter. The localization filters are learned by weighted binary cross entropy based on stride=4 maxpooling of pseudo masks. With the predicted maps, regions contain the catheter is selected as the input for the decoder. Although there are some literatures to adopt Mask R-CNN to extract the mask for segmentation decoder, it requires complex region proposal network with a large amount of training images. As a consequence, we adopted this straightforward and simple solution with the consideration of the efficiency and limited dataset \cite{huang20203}.

\begin{figure}[htbp]
\centering
\includegraphics[width=8cm]{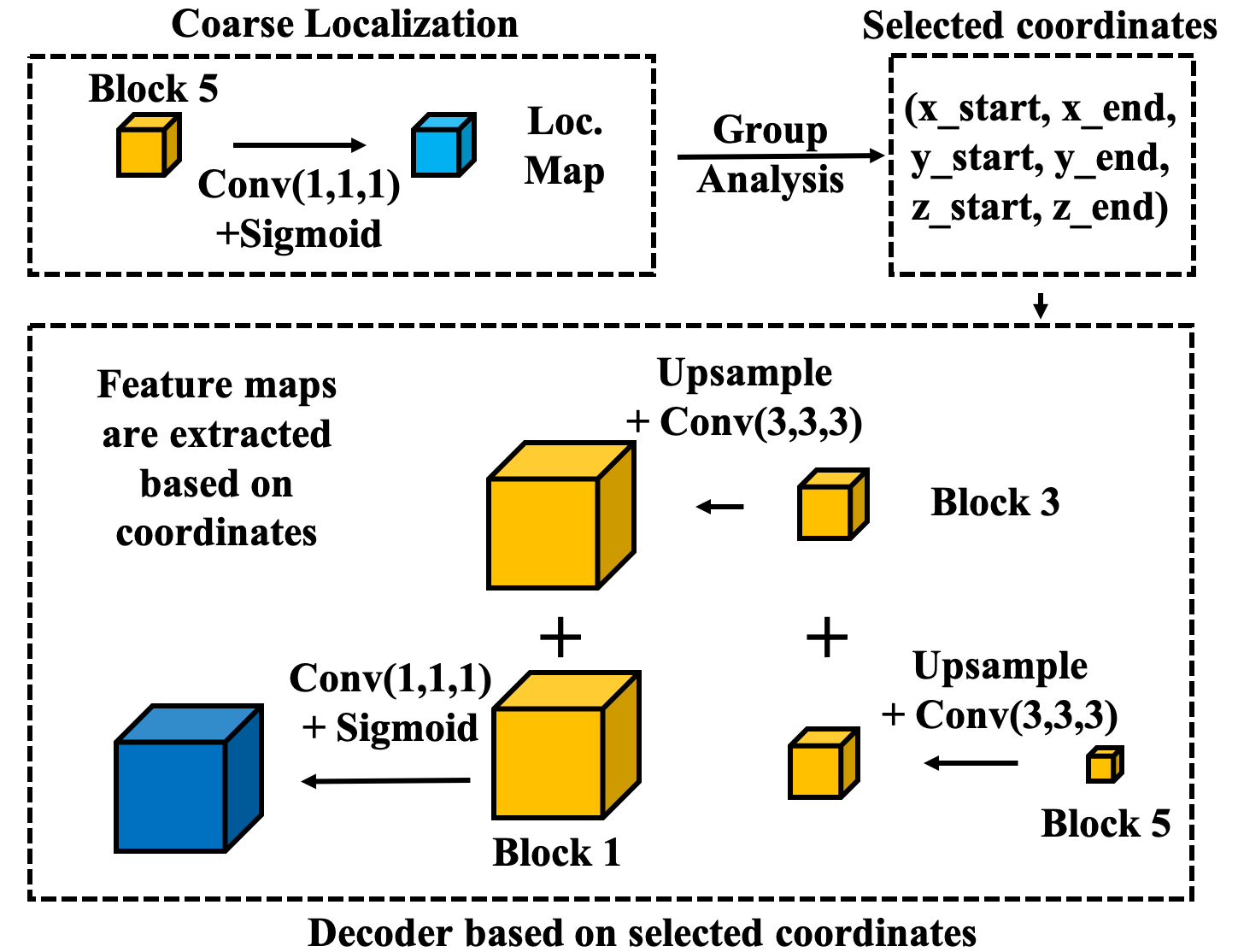}
\caption{Catheter localization and segmentation module based on the backbone ResNet encoder, which is trained by pseudo labels. Moreover, it is jointly optimized with intra-volume classification branch.}
\label{locseg}
\end{figure}

Based on the selected regions from localization map, i.e. spatial coordinate $(x_\text{start},x_\text{end},y_\text{start},y_\text{end},z_\text{start},z_\text{end})$ by grouping and region measurement, corresponding multi-scale feature maps are extracted from Block 1, 3 and 5 in ResNet-10 in Fig.~\ref{overivew}. Then, a simple decoder is introduced to obtain the segmentation results. Based on the backbone of the ResNet, feature maps of Block 5 is firstly upsampled to match the scale of Block 3, and then the convolution with kernel size 3 is applied to reduce the size of feature map to 64, which is concatenated to Block 3 with one extra convolution layer of kernel number 64. Similar operations are applied to Block 1, and finally one sigmoid is applied to obtain the segmented results. The decoder is trained based on the ROIs prediction and corresponding pseudo masks, while the final volume is reconstructed with known spatial coordinates.

\noindent {\bf Loss function and joint training}: Because of three different branches for different tasks, we design a joint loss function $L_\text{joint}$ for the network, which is defined as
\begin{equation}\label{TotalLoss}
L_\text{joint}=L_\text{cls}+L_\text{loc}+L_\text{seg},
\end{equation}
where $L_\text{cls}$ is a standard binary cross entropy (BCE) to train the classifier. $L_\text{loc}$ is a weighted binary cross entropy to address the unbalanced classes distribution for catheter localization. As for $L_\text{seg}$, a standard Dice + BCE loss is employed to train the segmentation decoder.

Because of the weakly supervised learning with different tasks, the end-to-end training is divided into three phases. First, the classifier is trained by $L_\text{cls}$. For each input volume, with bounding box annotation, $N$ regions containing catheter and $N$ regions without catheter are randomly sampled from the images. With the obtained coordinates, their corresponding feature maps after Block 5 are extracted for classification branches. The classifier is trained to learn the discriminative information of the catheter and is used to generate CAMs. Second, based on CAMs and line filtering responses, the pseudo masks are generated by dense CRF, which is then used to train location and segmentation branches. The location branch is trained together with $L_\text{cls}$ by sharing the same backbone encoder. Third, after location brach is converges, i.e. could correctly provide the coordinate of the catheter, $L_\text{joint}$ is optimized as a multi-task learning approaches to learn the segmentation decoder. During the training $M$ possible locations in localization maps are extracted for segmentation loss calculation.
\section{Experiments}\label{Experiments}
In this section, the dataset, implementation details and evaluation metrics a discussed.
\subsection{Dataset and preprocessing}
\begin{figure}[htbp]
\centering
\includegraphics[width=6cm]{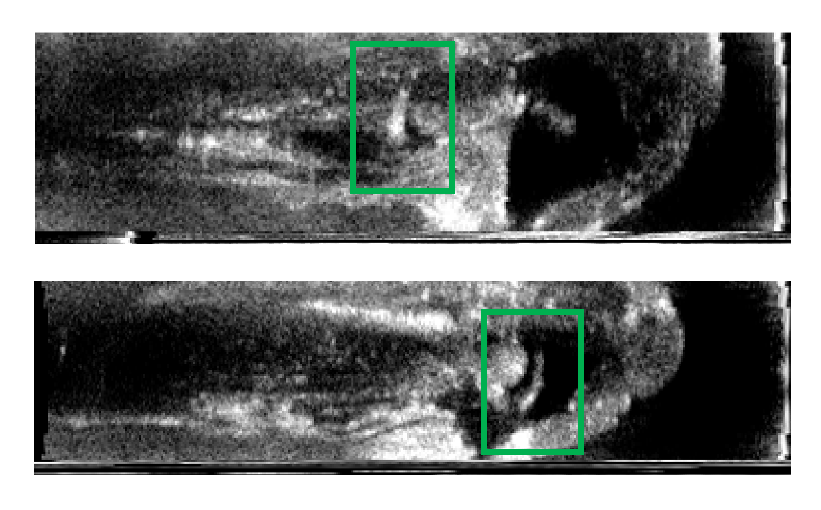}
\caption{Example slices of input volume and bounding box annotation (green). Top: training image, bottom: testing image.}
\label{imageexample}
\end{figure}

To validate our method, we collected an \emph{ex-vivo} dataset on RF-ablation catheter for cardiac intervention, consisting of 60 3D cardiac US volumetric data from three isolated porcine hearts. During the recording, the heart was placed in a water tank with the RF-ablation catheter (with the diameter of 3.3 mm) inside the heart chambers. The phase-array US probe (X7-2t with 2,500 elements by Philips Medical Systems, Best, Netherlands) was placed next to the interested chamber to capture the images containing the catheter, which was monitored by a US console (EPIQ 7 by Philips, high-resolution mode). For each recording, we pull out the catheter and re-inserted it into the heart chamber to reduce the dependence among images, i.e., one image for one session. Meanwhile, the US probe was tuned to have different capturing view. All the volumes were manually annotated at voxel level in Frustum image. Moreover, bounding box contains catheter is also annotated w.r.t location and length in 3D volume for method validation. Specifically, we collected 45 training images in the ventricles. In contrast, 15 volumes were collected in the right atrium, which is then divided into 5 and 10 volumes as validation and testing datasets. Therefore, the training and testing images have no information overlapping. The volume size is around $360\times96\times96$ for the data in the Frustum domain. Corresponding voxel units are $0.2695 \text{mm}\times1.003^{\circ}\times1.003^{\circ}$. Example slices from volumes are shown in Figure.~\ref{imageexample}. As can be observed, anatomical structures are different between two different datasets. Moreover, the bounding box we obtained is relatively loose when compared to the actual size of the catheter.

\subsection{Implementation details}
We implemented our framework in Python 3.7 with PyTorch 1.1.0, using a standard PC with a TITAN 1080Ti GPU. To construct the network, we use ResNet10 as our backbone encoder, which employs the pre-trained parameters from Med3D project \cite{chen2019med3d,he2016deep}. During the training, classification branch is firstly trained with 100 epochs to generate the CAMs with $N=16$ (since the global average pooling is hard to converge than a standard FC layer), after which 30 epochs training for localization branch is applied (we experientially weighted positive voxels are 10 times important than background voxels). Finally, jointly training with 20 epochs is firstly applied to confirm the network convergency, then the iteration label updating is applied until validation Dice score is converged or reached 50 epoch in total. Learning rate is set to be 0.0001 for all parameters, which are optimized by Amsgrad optimizer \cite{reddi2019convergence}. Moreover, mini-batch is set to be one due to limited GPU memory. Pseudo label generation is performed on-the-fly for every 4 epochs for jointly training, which refines the pseudo label based on Eqn.~(\ref{update}). Moreover, $N=16$ random samples are generated around the Bbox to obtain a smooth prediction of CAMs. Meanwhile, $M=10$ for training phase and $M=2$ for testing step. As parameters for the pairwise interactions in dense CRF, we experientially optimized them based on the authors’ public implementation \cite{krahenbuhl2011efficient,kamnitsas2017efficient}. We excluded the CRF inference in testing stage as it is computational expensive with long time processing.

\subsection{Evaluation metrics}
To evaluate the overall segmentation performance of the proposed method, we consider the Dice score (DSC) and volumetric similarity (VS) as evaluation metrics \cite{taha2015metrics}. As for DSC, it is defined as: 
\begin{equation}\label{DSC}
\text{DSC}=\frac{\text{2TP}}{\text{2TP+FP+FN}},
\end{equation}
where the TP denotes true positive, FP is false positive and FN is false negative. The Volume Similarity or VS value is defined by 
\begin{equation}\label{VS}
\text{VS}=1-\frac{|\text{FN}-\text{FP}|}{\text{2TP+FP+FN}}.
\end{equation}
It is an overlap-based metric, since the absolute volume of the segmented region is compared with the corresponding ground truth. This metric is useful to evaluate whether the segmentation has a volume as close to the ground truth as possible \cite{taha2015metrics}. 

\section{Results}\label{Results}
In this section, several comparison experimental results are shown, including ablation studies and comparisons to other segmentation methods. All the results are evaluated based on the 10 testing images.
\subsubsection{Ablation studies}
In this part, several ablation studies are performed to evaluate the effectiveness of weakly supervised learning framework. Specifically, three different ablation studies were performed to evaluate the different components of our training scheme.

\noindent {\bf Pseudo label generation}: In this ablation study, we evaluate different approaches for pseudo label generation: (1) Purely bounding box based annotation without any refinement as the baseline for the weakly supervised segmentation, which is denoted as Bbox. (2) The Bbox overlaps with CAMs is considered as the second approach, which is denoted as Bbox+CAMs. (3) Based on Bbox, line filtered results are overlaid on top of Bbox to generate a label, which is denoted as Bbox+LF. (4) Based on Bbox+CAMs, Frangi line filtering is included as a further refinement for the psuedo boundary, denoted as Bbox+CAMs+LF. (5) the proposed pseudo label generation method as described in previous sections without pseudo label updating, which is denoted as Bbox+CAMs+LF+CRF, and finally (6) The proposed method with pseudo label updating. The results are shown in Table~\ref{TB1}. 

\begin{table}[htbp]
\centering
\caption{Ablation study on different approaches for pseudo label generation, which are are evaluated by Dice score (DSC) and volumetric similarity (VS) using mean$\pm$std.}
\label{TB1}
\begin{tabular}{l|c|c}
\hline
Methods&DSC (\%)&VS (voxel)\\ \hline
Bbox &14.6$\pm$10.3&22.4$\pm$22.1\\\hline
Bbox+CAMs &47.9$\pm$13.2&53.9$\pm$17.7.6\\\hline
Bbox+LF&33.9$\pm$15.3&37.6$\pm$18.8\\\hline
Bbox+CAMs+LF &54.7$\pm$30.3&72.5$\pm$17.3\\\hline
Bbox+CAMs+LF+CRF&62.9$\pm$23.3&74.5$\pm$16.1\\\hline\hline
Proposed &66.6$\pm$10.3&83.2$\pm$14.8\\\hline
\end{tabular}
\end{table}

As can be observed, the proposed pseudo label generation has much better performance than other methods. Since the proposed method provides a refined and accurate catheter information iteratively during the training updating. More specifically, the combination of CAMs+LF provides a richer catheter related information based on CAMs-guided spatial attention and shape priori. With a further CRF refinement, a better pseudo label is obtained for training. Comparing the proposed method to the one without label updating, the updating strategy provides better performance, which is because of the iteratively learned information from segmentation network itself, and adopt it for discriminative information extraction. It is worth to mention that we failed to obtain a successful result based on line filtered images only as the pseudo label, since it includes too much outliers to segment the target catheter.

\noindent {\bf Training procedures}: Different training procedures are also compared to show the effectiveness of the joint training for different branches and transfer learning. (1) The randomly initialized network that three branches are trained separately at different stage without joint optimization by Eqn.~(\ref{TotalLoss}). (2) The condition of three branches are trained separately at different stage without joint optimization by Eqn.~(\ref{TotalLoss}) but with pre-trained ResNet-10 backbone from Med3D project \cite{chen2019med3d}. (3) The proposed weakly supervised learning method without considering the pre-trained ResNet-10. parameters. (4) The scheme that the proposed joint training for three different branches based on pre-trained ResNet-10. For cases without pre-trained initialization, the learning rate is set to be 0.001. The results are shown in Table~\ref{TB2}. 

\begin{table}[htbp]
\centering
\caption{Ablation study on different training scheme, which are are evaluated by Dice score (DSC) and volumetric similarity (VS) using mean$\pm$std.}
\label{TB2}
\begin{tabular}{l|c|c}
\hline
Methods&DSC (\%)&VS (voxel)\\ \hline
Train separately w/o pre-train &49.1$\pm$27.8&63.0$\pm$18.4\\\hline
Train separately w pre-train &58.8$\pm$19.3&68.9$\pm$24.8\\\hline
Train jointly w/o pre-train &54.9$\pm$11.6&71.3$\pm$16.5\\\hline
Train jointly w pre-train&66.6$\pm$10.3&83.2$\pm$14.8\\\hline
\end{tabular}
\end{table}

From the table, joint the train provides a better performance than separate training. As we observed during the training procedure, separate training would lead to a worse classification performance as the segmentation network goes to converge, therefore a worse pseudo labels are generated due to this limitation. In contrast, joint training constraint the network as a multi-task learning that a better performance is achieved. By comparing cases with and without pre-trained parameter, employing pre-trained parameters can increase the segmentation performance than train from scratch by a large margin. These results indicate the pre-train model could boost the segmentation result with limited training data.
\subsubsection{Comparison to other methods}
Although the weakly supervised segmentation has been studied in recent years, it still not widely studied in 3D medical imaging area. Wang \emph{et al.} \cite{wang2020iterative} proposed an iterative denoising network for bounding box based weakly supervised learning. Nevertheless, without open sourced code, we failed to implement their method with successful results. Alternatively, we considered GrabCut-Itr \cite{rother2004grabcut} to refine the bounding box. As is shown by Wang \emph{et al.} \cite{wang2020iterative}, this strategy is also working on their tasks. Specifically, the GrabCut segmentation is generated based on the super voxels from Frangi line filtered results. Then, the segmented GrabCut is used to train the localization and segmentation network of our framework. Similar to the proposed method, the GrabCut segmentation is iteratively updated based on time average moving, which gradually refines the pseudo annotation. Furthermore, we also performed comparison to the experiments in Cartesian domain. Specifically, the considered datasets are scan converted based on Frustum domain, which have image size of $360\times360\times336$ voxels (around $0.2^3$ mm$^3$ for each voxel based on QLab, Philips). Nevertheless, a common GPU memory cannot handle the huge size of the images with complex 3D operations. To address this, a compromised approach is applied to reduce model size: for each layer, number of filters are downsampled by a factor of 2 based on original Med3D implementation \cite{chen2019med3d} to fit GPU memory (stride=8 for final feature map size). Finally, a fully supervised learning approach based on our training data is also applied to define the upper bound of the proposed segmentation network. It is worth to mention that although peak response map based faster r-cnn \cite{dong2019instance} was proposed for bounding box based weakly segmentation, we cannot reproduce the results based on their method. Since the faster r-cnn network is failed to train on our imbalanced and limited training dataset, which have extremely large amount of background voxels. Moreover, the method of \cite{dong2019instance} assumes the target as a ball shape for peak response map generation, which cannot fit our task.
\begin{table}[htbp]
\centering
\caption{Comparison to other methods, which are are evaluated by Dice score (DSC) and volumetric similarity (VS) using mean$\pm$std. }
\label{TB3}
\begin{tabular}{l|c|c}
\hline
Methods&DSC (\%)&VS (voxel)\\ \hline
GrabCut-Itr &20.5$\pm$20.7&63.1$\pm$30.0\\\hline
Cartesian&44.7$\pm$25.3&70.8$\pm$19.4\\\hline
Fully supervised &70.0$\pm$9.2&81.4$\pm$12.4\\\hline\hline
Proposed &66.6$\pm$10.3&83.2$\pm$14.8\\\hline
\end{tabular}
\end{table}

\begin{figure*}[htbp]
\centering
\includegraphics[width=12cm]{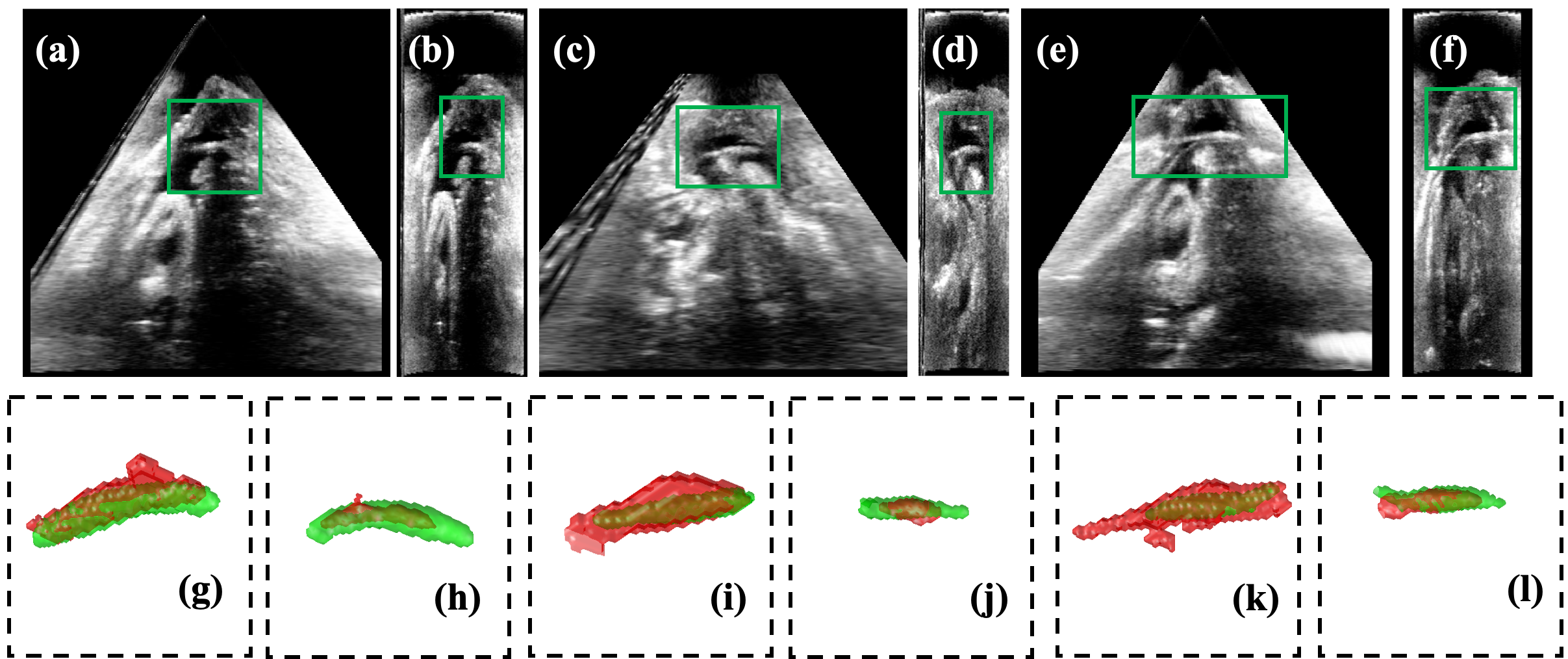}
\caption{Example images and corresponding segmentation results. Cartesian images: (a), (c), (e), Cartesian segmentation (g), (i) (k). Frustum images: (b), (d), (f), Frustum segmentation (h), (j) (l). Green: voxel-level annotation or bounding box, red: segmentation results. }
\label{Cartesian}
\end{figure*}

\begin{figure*}[htbp]
\centering
\includegraphics[width=16cm]{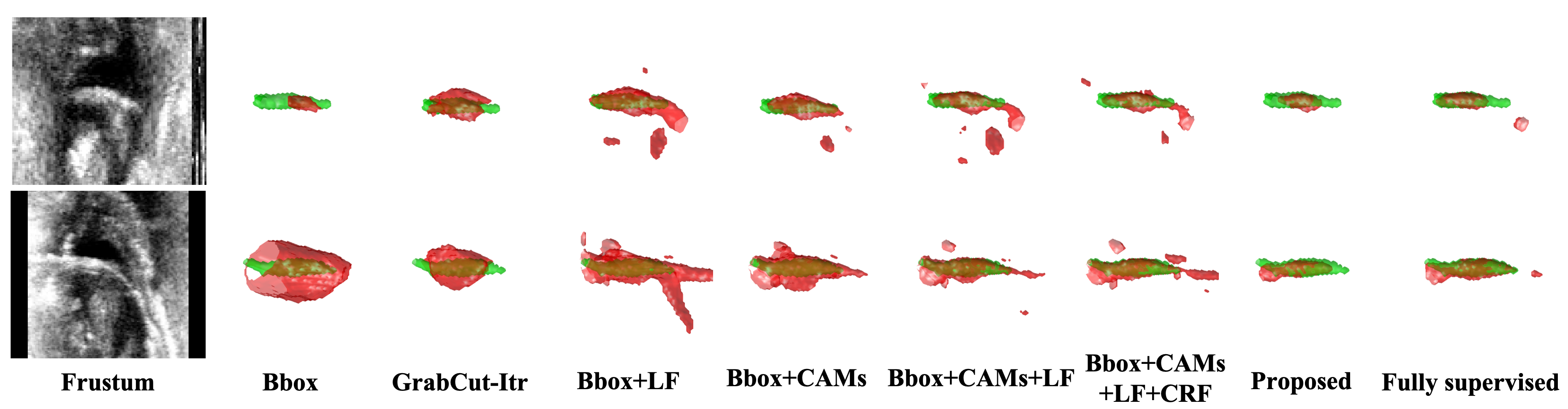}
\caption{Example segmentation results for different methods in Table~\ref{TB2} and Table~\ref{TB3}. Green: voxel-level annotation, red: segmentation results.}
\label{Segments}
\end{figure*}

From the the Table~\ref{TB3}, the GrabCut provides worse performance than the proposed method, as the initialized the pseudo annotation is not good enough to guide the segmentation. Although the iterative updating is applied to refine the annotation, it is still worse than line filter based pseudo label, which is because of a better shape description based on the information combination from CAMs and CRF. Compared to Cartesian images with same framework, our method outperform it by a large margin. There are two main reasons: (1) The Cartesian image is too large to train, which is compromised by a simplified network model. As a result, the pre-trained parameter is not available for our limited dataset. In contrast, as is shown in Table.~\ref{TB2}, transfer learning could better address this limitation. (2) The Cartesian image is too large to be learned, as it contains more data points than a Frustum image. However, Cartesian image less information than Frustum ultrasound from theoretical view. Although we obtained similar inference time for Cartesian and Frustum images, which are around 0.25 seconds, but this is not a fair comparison as the network architectures are different, which have different image stride size and number of filters. Compared to fully supervised learning method, our proposed method obtains a worse performance in Dice score. Nevertheless, with respect to volumetric similarity, their results are similar, which indicates these methods have similar output shape. 

Example images of segmentation results are shown in Figure.~\ref{Cartesian} and Figure.~\ref{Segments}. The comparisons between Cartesian and Frustum domains are shown in Figure.~\ref{Cartesian}. From the result, the Cartesian image has worse performance with outliers and false positives. From the training, we observed the learning curve is much harder to converge than transfer learning, which is a key reason to receive this worse performance. Comparisons among different methods, i.e. ablation studies and comparison to other methods, are shown in Figure.~\ref{Segments}. As can be observed, Bbox based annotation cannot generate meaningful semantic results, which is similar to GrabCut. Alternatively, line filtering and CAMs based methods could extract catheter orientation information and also boundary details. Nevertheless, compared to their combination, they are less discriminative for detailed information. Moreover, CRF processing during annotation generation shows similar appearance to the case without CRF. However, with iterative updating of the pseudo label, the outliers can be removed by the gradually refinement. By comparing the proposed method to fully supervised results, their visual results have differences that fully supervised learning tends to have higher false positive rate with higher positive voxels prediction.
\begin{table}[htbp]
\centering
\caption{Computation complexity analysis by GFLOPs. }
\label{TB4}
\begin{tabular}{l|c|c}
\hline
Data format&w/o ROI-Seg&w ROI-Seg\\ \hline\hline
Cartesian&68.2&23.9\\\hline
Frustum&5.2&1.8\\\hline\hline
\end{tabular}
\end{table}

As for the computation complexity, we performed FLOPs \cite{molchanov2016pruning} analysis on the proposed method in four different situations. The results are shown in Table~\ref{TB4}. As can be observed, the proposed model with ROI-Seg achieves the lowest computational cost in Frustum US, while it has higher cost when applying the segmentation on the whole volume (we failed to train the model due to out of memory). In contrast, its corresponding Cartesian US would leads to a higher computation cost, meanwhile, the proposed model cannot be loaded with a standard GPU memory.
\subsection{Discussion}
The proposed method achieved the state-of-the-art segmentation performance by weakly supervised learning, yet more crucially less cost of GPU memory and computations by employing 3D Frustum US. However, to apply the proposed method for real-time clinical applications, there still few discussion aspects for it. First, the pseudo annotation is generated based on CRF inference with probability mapping, which is however may not exploit the 3D spatial information properly end-to-end parameter optimization. Although there are some works to optimize the 3D correlations by deep learning, such as ConvCRF \cite{teichmann2018convolutional}, it is expensive to use it in a 3D volumetric data by using a standard GPU. As a result, a cheaper and convenient optimization method should be considered for CRF inference. Second, a lower resolution mode of US machine should be exploited in future, as it would to smaller volume size for Frustum images, which yield a faster inference time. Nevertheless, a lower image resolution might degrade the segmentation performance for clinical usage. Third, further validation on \emph{in-vivo} dataset is required to support the clinical value for the proposed method, which is considered as future work. Specifically, the employed dataset has different appearance and image qualities than the clinical practice. Since the US images were collected in a water tank with filled water, which is however should be blood for a human heart. This difference could lead to different image noise and contrast, which might increase the challenges for the real-clinical tasks. Fourth, the processed Frustum domain does not follow the standard Euclidean space, which may degrade the performance without proper spatial measurement for training. Point cloud based segmentation with Riemannian space might be exploited for a better performance. Finally, the proposed segmentation method exploit the static images instead of 3D video, which is commonly obtained by the operation. Therefore, the temporal information should be exploited in the future.
\section{Conclusions}\label{Conclusions}
3D US-guided therapy has been widely studied, but it is difficult for surgeons to localize the instrument in 3D volumetric data, because of complex anatomical structures in the tissue. Therefore, automated instrument segmentation, such as catheter segmentation, is essential to reduce the operation effort and increase the outcome. Nevertheless, the existing deep learning method is not feasible for clinical practice due to expensive image annotation and huge size of 3D volumetric data. In this article, we propose a novel weakly supervised learning method for catheter segmentation in Frustum US image, which avoids expensive voxel-level annotation by incorporating the prior knowledge of the catheter. Moreover, with 3D Frustum image, the computational requirements are drastically reduced when compared to a conventional Cartesian US images. With the obtained state-of-the-art performance, our proposed method paves the way for real-time clinical applications with less annotation effort.
\bibliographystyle{IEEEtran}
\bibliography{draft}
\end{document}